\documentclass{bmvc2k}

\title{CVAM-Pose: Conditional Variational Autoencoder for Multi-Object Monocular \\ Pose Estimation}

\addauthor{Jianyu Zhao}{jzhao12@uclan.ac.uk}{1}
\addauthor{Wei Quan}{wquan@uclan.ac.uk}{1}
\addauthor{Bogdan J. Matuszewski}{bmatuszewski1@uclan.ac.uk}{1}

\addinstitution{
 Computer Vision and Machine Learning (CVML) Group,\\
 The University of Central Lancashire,\\
 Preston, UK
}

\runninghead{Jianyu, Wei, Bogdan}{CVAM-Pose}

\usepackage{hyperref}
\usepackage{amssymb}
\usepackage{amsmath}
\usepackage{enumitem}
\usepackage{multirow}
\usepackage{bm}

\begin{document}

\maketitle

\begin{abstract}

Estimating rigid objects' poses is one of the fundamental problems in computer vision, with a range of applications across automation and augmented reality. Most existing approaches adopt one network per object class strategy, depend heavily on objects’ 3D models, depth data, and employ a time-consuming iterative refinement, which could be impractical for some applications. This paper presents a novel approach, CVAM-Pose, for multi-object monocular pose estimation that addresses these limitations. The CVAM-Pose method employs a label-embedded conditional variational autoencoder network, to implicitly abstract regularised representations of multiple objects in a single low-dimensional latent space. This autoencoding process uses only images captured by a projective camera and is robust to objects' occlusion and scene clutter. The classes of objects are one-hot encoded and embedded throughout the network. The proposed label-embedded pose regression strategy interprets the learnt latent space representations utilising continuous pose representations. Ablation tests and systematic evaluations demonstrate the scalability and efficiency of the CVAM-Pose method for multi-object scenarios. The proposed CVAM-Pose outperforms competing latent space approaches. For example, it is respectively 25\% and 20\% better than AAE and Multi-Path methods, when evaluated using the $\mathrm{AR_{VSD}}$ metric on the Linemod-Occluded dataset. It also achieves results somewhat comparable to methods reliant on 3D models reported in BOP challenges. Code available: \url{https://github.com/JZhao12/CVAM-Pose}

\end{abstract}

\section{Introduction}
\label{sec:intro}

The rapid and precise estimation of rigid objects' poses with six degrees of freedom (6-DoF) is crucial for a wide range of real-world applications, including explorative navigation, augmented reality, and automated medical intervention. The introduction of deep learning techniques~\cite{Goodfellow2016,zhang2023dive} marked a significant evolution in computer vision, yielding remarkable outcomes in 6-DoF pose estimation. Notably, most deep learning-based methods~\cite{xiang2018posecnn,wu2018real,li2020deepim,park2019pix2pose,li2019cdpn,peng2020pvnet,song2020hybridpose,wang2021gdr,sundermeyer2020augmented,deng2020self,deng2021poserbpf,zhao2023cvml,su2022zebrapose} tend to train individual networks for each object to obtain higher pose accuracy. However, these approaches are resource-consuming compared to training a unified multi-object network, as memory usage increases with the number of objects (networks). Additionally, most methods typically require 3D models~\cite{xiang2018posecnn,wu2018real,bukschat2020efficientpose,labbe2020cosypose,li2020deepim} or establish 2D-3D correspondences based on the models~\cite{zakharov2019dpod,park2019pix2pose,hodan2020epos,li2019cdpn,peng2020pvnet,rad2017bb8,tekin2018real,song2020hybridpose,su2022zebrapose}, where the need for 3D models can be seen as one of the limiting factors for broader applications.

\begin{figure}[ht]
\centering
\includegraphics[width=\textwidth]{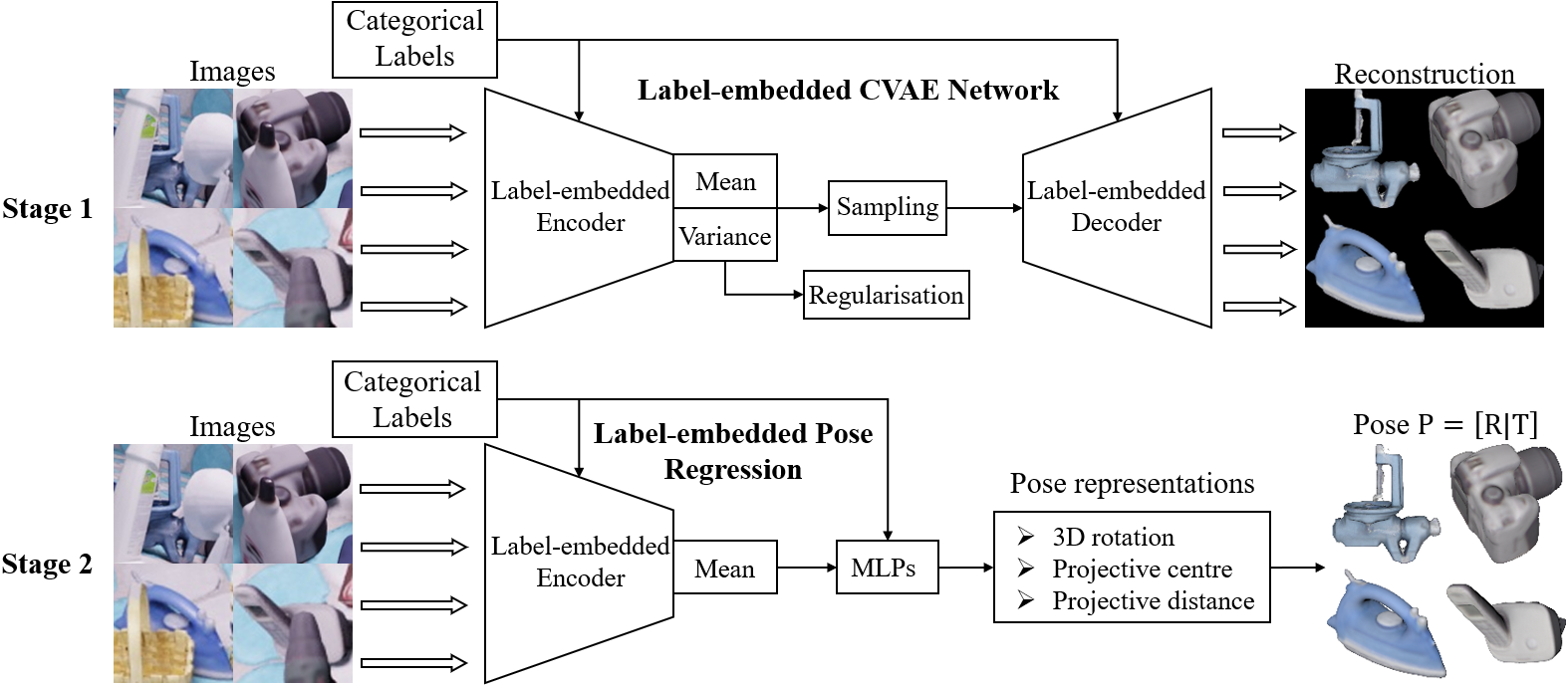}
\caption{During training, the label-embedded CVAE network abstracts information from both images of objects and the corresponding categorical labels in the latent space, which are then interpolated to multi-object 6-DoF poses using MLPs. The images of objects are taken from the Linemod PBR dataset~\cite{hinterstoisser2013model,hodavn2019photorealistic,hodavn2020bop,pbrdata}.}
\label{fig:cvae6d pipeline}
\end{figure}

In this paper, a novel multi-object pose estimation method called CVAM-Pose is proposed (Fig.~\ref{fig:cvae6d pipeline}), which contains two main stages. The first stage involves training a label-embedded conditional variational autoencoder (CVAE) network that incorporates the one-hot encoding technique to facilitate the learning of regularised and constrained representations of multi-object poses in the latent space. Different from the original CVAE network proposed in~\cite{sohn2015learning}, the adapted layer-wise one-hot encoding technique encodes categorical labels as complete feature maps across every layer within the network, enhancing the learning of high-level representations. The second stage applies label-embedded pose regression that avoids the discretisation of poses. This involves concatenating the learnt multi-object representations with one-hot encoded label vectors, and training multilayer perceptrons (MLPs) to regress these concatenated features into continuous pose representations. The contributions of the CVAM-Pose method are summarised as follows:

\begin{enumerate}
  \item The method enhances the scalability and efficiency for multi-object pose estimation using a single CVAE network. To the best of our knowledge, it is the first time a conditional generative model is employed to efficiently characterise multi-object poses. The adapted label-embedding technique also improves the capability of learning high-level representations.
  \item The method does not require object 3D models, depth data, and post-refinement during inference, which can facilitate real-time processing. It achieves results comparable to the state-of-the-art approaches on the Linemod-Occluded benchmark dataset and outperforms those based on latent space representation.
  \item The method effectively addresses various challenging scenarios, including texture-less objects, occlusion, truncation~\cite{peng2020pvnet}, and clutter.
\end{enumerate}

\section{Related Work}
\label{sec:related}

\noindent\textbf{Deep Learning-based Approaches}\hspace{10mm}With the rapid development of deep learning techniques, numerous state-of-the-art pose estimation methods employing convolutional neural networks (CNNs) have been proposed. These methods can be categorised into three distinct groups based on their approach to utilising CNNs: direct, indirect, and latent representation methods.

The direct methods train CNNs to regress 3D rotation and translation from images directly. which either construct loss functions using 3D model points~\cite{xiang2018posecnn,wu2018real,bukschat2020efficientpose}, or iteratively match the image rendered from a 3D model at its estimated pose with the observed input image~\cite{labbe2020cosypose,li2020deepim}. Typically, these methods reparameterise rotation into representations more suitable for network training, such as unit quaternion~\cite{dam1998quaternions}, axis-angle~\cite{tomasi2013vector}, or continuous representation~\cite{zhou2019continuity}. The indirect methods focus on learning 2D-3D model correspondences via CNNs, with the 6-DoF poses subsequently estimated using PnP~\cite{gao2003complete,lepetit2009epnp} and RANSAC~\cite{fischler1981random}. The model correspondences can be in the form of pixel-wise dense mapping~\cite{zakharov2019dpod,park2019pix2pose,hodan2020epos,li2019cdpn,song2020hybridpose,haugaard2022surfemb,su2022zebrapose}, or a selection of sparse keypoints~\cite{peng2020pvnet,rad2017bb8,tekin2018real}. The latent representation methods learn implicit latent space representations using specific network architectures, typically autoencoders. The pose of a test instance is often retrieved using a lookup table (LUT) technique, which includes finding nearest neighbours~\cite{sundermeyer2020augmented,sundermeyer2020multi} and computing observation likelihoods~\cite{deng2020self,deng2021poserbpf}.

Both direct and indirect methods explicitly require accurate 3D models for training CNNs or establishing 2D-3D correspondences. The latent representation methods, despite using only images from single perspective camera, often suffer from the pose discretisation problem due to the nature of LUT.

\vspace{12pt}

\noindent\textbf{Conditional Variational Autoencoder}\hspace{10mm}The variational autoencoder (VAE)~\cite{kingma2013auto,kingma2019introduction} was introduced in the context of generative models, which is different from typical autoencoder models~\cite{hinton2006reducing,ranzato2007sparse,rifai2011contractive,vincent2008extracting}. The primary objective of the VAE is to generate new, typically highly dimensional data points, with the generation process controlled by a low-dimensional latent code randomly drawn from a prior distribution, such as Gaussian. However, a notable limitation is its inability to specify the characteristics of the generated data. To address this issue,~\citet{sohn2015learning} introduced the conditional variational autoencoder (CVAE), which extends the VAE framework to incorporate conditional parameters, thereby enabling the generation of data with desired attributes.

In the context of object 6-DoF pose estimation,~\citet{zhao2023cvml} proposed a VAE-based method called CVML-Pose, which was restricted to single-object predictions. Similar methods, such as~\cite{tatemichi2024category}, have also been developed, but using RGB-D images as input. We extend the CVML-Pose method by training a CVAE model with a layer-wise one-hot encoding technique. This adaptation facilitates the learning of multi-object representations in a single latent space, significantly improving scalability and efficiency in the prediction of multi-object poses.

\section{Methodology}
\label{sec:methodology}

\subsection{Implicit Learning of Multi-Object Representations}
\label{subsec:learning}

To effectively learn multi-object representations, a label-embedded CVAE network is trained to encode images of objects $x_i$ and their corresponding one-hot encoded categories $y_i$ as label conditions in a regularised latent space, subsequently outputting clean reconstructions $x_i^\prime$.

\begin{figure}[ht]
\centering
\includegraphics[width=\textwidth]{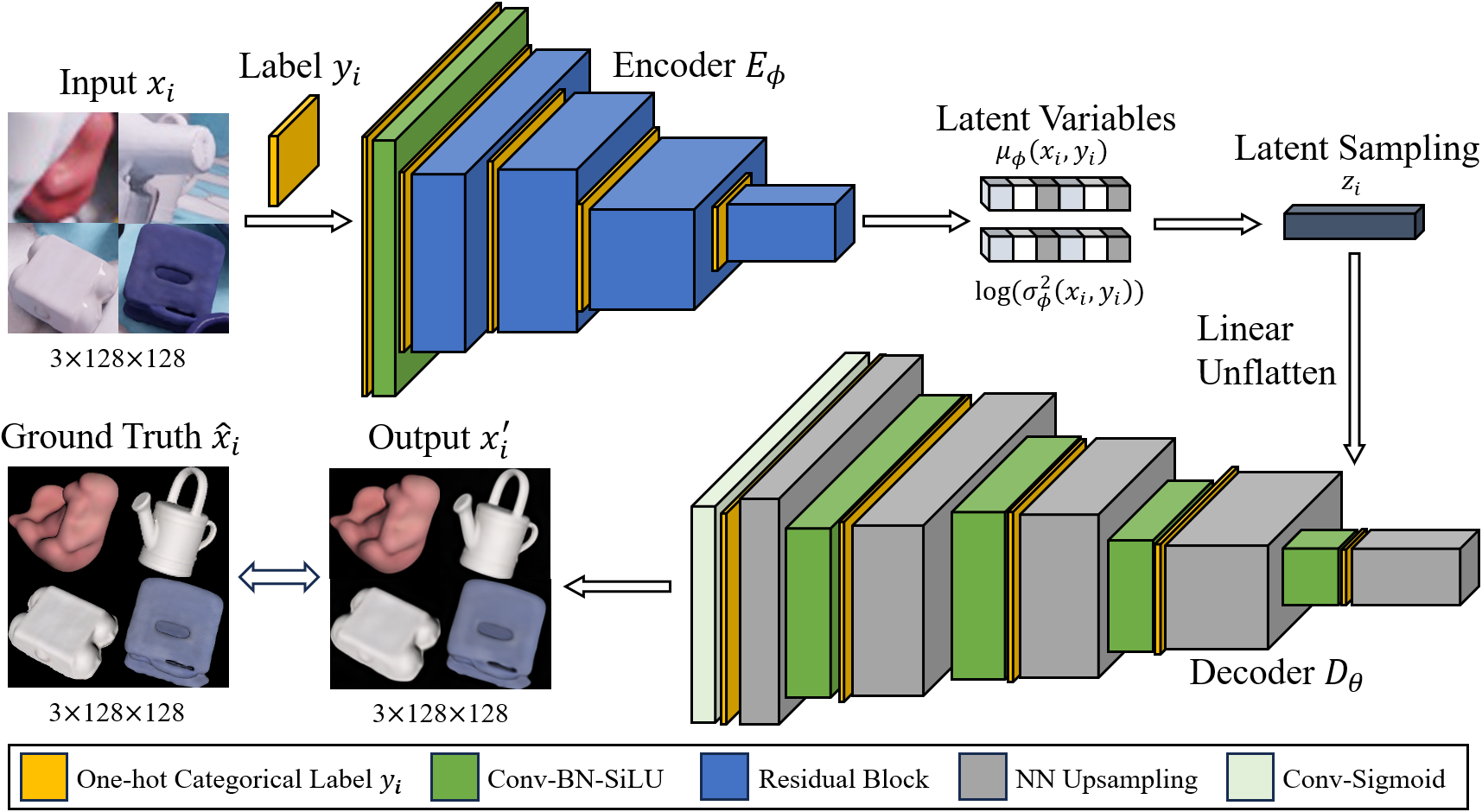}
\caption{The proposed label-embedded conditional variational autoencoder network. The images of objects are taken from the Linemod PBR dataset~\cite{hinterstoisser2013model,hodavn2019photorealistic,hodavn2020bop,pbrdata}.}
\label{fig:cvae6d autoencoder network}
\end{figure}

As depicted in Fig.~\ref{fig:cvae6d autoencoder network}, an asymmetric architecture is proposed, consisting of an encoder network $E_\phi$ and a decoder network $D_\theta$, with learnable parameters $\phi$ and $\theta$ respectively. The encoder $E_\phi(x_i,y_i)$ processes both $x_i$ and $y_i$, where $y_i$ are embedded as complete feature maps in every convolution block (block-wise), until the latent variables are obtained in the latent space, including $\mu_\phi(x_i,y_i) \in \mathbb{R}^n$ and $\log{(\sigma^2_\phi(x_i,y_i))} \in \mathbb{R}^n$, where $(\mu_\phi,\sigma^2_\phi)$ represent mean and variance vectors of the multivariate normal distribution. Due to the non-differentiability of sampling from $\mathcal{N}(z_i;\mu_\phi(x_i,y_i),diag(\sigma_\phi(x_i,y_i)))$, a reparameterisation trick~\cite{kingma2013auto} is employed. The latent sampling $z_i \in \mathbb{R}^n$ is reparameterised as $\mu_\phi(x_i,y_i) + diag(\sigma_\phi(x_i,y_i))\cdot\epsilon$, where $\epsilon\sim \mathcal{N}(0,I)$. After sampling, the decoder network $D_\theta(z_i,y_i)$ reconstructs the complete and clean view $x_i^\prime$ from both $z_i$ and $y_i$, where $y_i$ is also embedded in every convolution layer (layer-wise) in the decoder.

For network training, the evidence lower bound $(ELBO)$ loss~\cite{zhao2023cvml} is used, assuming a Gaussian prior distribution $p(z)=\mathcal{N}(z;0,I)$. This loss comprises two components: i) a pixel-wise squared L2 norm between the output image $x_i^\prime$ and the ground truth reconstruction image $\hat x_i$; ii) a Kullback-Leibler (KL) divergence loss with a scalar $\alpha$, which controls the regularisation of the latent space.
\begin{equation}\label{eq:ELBO}
ELBO \simeq - \sum_{i=1}^{m}\Biggl({||\hat x_i-x_i^\prime||}^2 - \alpha \cdot \sum_{j=1}^{n}{\Bigl(1+\log{({\sigma^2_{ij}})}-{\mu^2_{ij}}-{\sigma^2_{ij}}}\Bigl)\Biggl)
\end{equation}
where $\mu_{ij}$ refers to the $j$-th element of the vector $\mu_i$, $\sigma^2_{ij}$ refers to the $j$-th element of the vector $\sigma^2_i$, $\mu_i=\mu_\phi(x_i,y_i)$, $\sigma^2_i=\sigma^2_\phi(x_i,y_i)$, $m$ represents the number of training data, and $n$ is the dimensionality of the latent space.

\begin{figure}[ht]
\centering
\includegraphics[width=\textwidth]{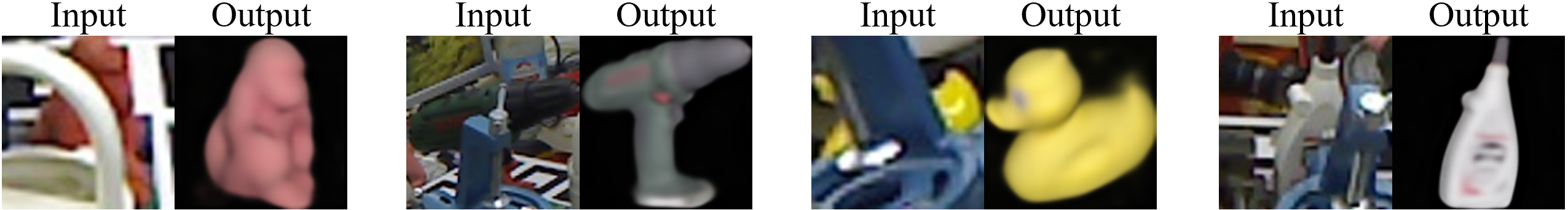}
\caption{The output images from decoder show objects' representations with occlusion and clutter removed. The test input images also shown are taken from the Linemod-Occluded dataset~\cite{brachmann2014learning,lmodata}.}
\label{fig:cvml recon example}
\end{figure}

After training, the label-embedded CVAE network has learnt robust representations of objects, possibly including poses. This can be evidenced by the clean and complete reconstructions produced from the trained network based on the test data. As illustrated in Fig.~\ref{fig:cvml recon example}, these reconstructions not only preserve complete views of objects but also diminish irreverent information such as occlusion and cluttered background.

\subsection{Continuous Regression for Multi-Object Pose Estimation}
\label{subsec:continuous}

\begin{figure}[ht]
\centering
\includegraphics[width=\textwidth]{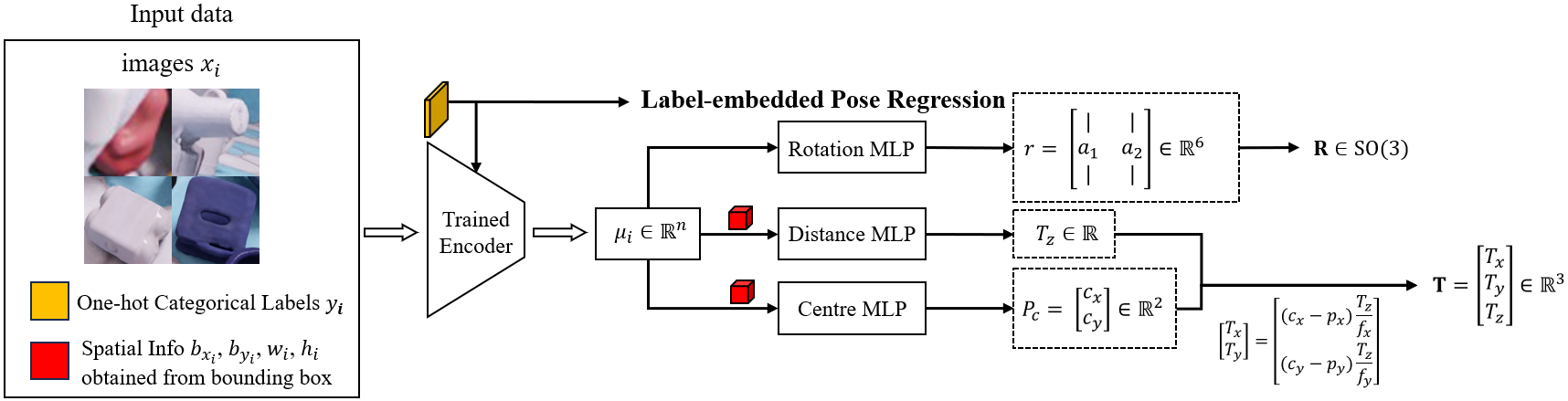}
\caption{The proposed label-embedded pose regression approach interpolates multi-object representations to continuous pose representations using multiple MLP heads.}
\label{fig:cvae6d pose regression}
\end{figure}

The subsequent stage employs a continuous pose regression strategy~\cite{zhao2023cvml}, with an adaptation to handle the multi-object scenario. As detailed in Fig.~\ref{fig:cvae6d pose regression}, this strategy utilises one-hot encoded object labels $y_i$ to train MLPs, enabling smooth interpolation of poses.

For estimating 3D rotation, the rotation MLP is trained to regress from $(\mu_i,y_i)$ to the continuous 6D rotation representation $r\in\mathbb{R}^6$~\cite{zhou2019continuity}, and the output rotation $\mathbf{R}\in \mathrm{SO}(3)$ is derived from $r$ through a process similar to Gram-Schmidt orthogonalisation. This continuous representation has proven more effective than others such as unit quaternion and axis-angle, and has been successfully implemented in~\cite{labbe2020cosypose,wang2021gdr}.

The estimation of 3D translation $\mathbf{T}=\left(\begin{matrix}T_x&T_y&T_z\\\end{matrix}\right)^T\in\mathbb{R}^3$ is disentangled into estimating the 2D projective centre coordinates $P_c=\left(c_x,c_y\right)^T\in\mathbb{R}^2$ and the projective distance $T_z\in\mathbb{R}$. Specifically, the latent vector $\mu_{i}$ is concatenated with spatial information obtained from the object's bounding box, including its width $w_i$, height $h_i$, and top-left corner coordinates $P_{bbox}=\left(b_{x_i},b_{y_i}\right)^T$, as well the label $y_i$. The dedicated MLP is then trained to regress these concatenated features to $\left(c_x,c_y\right)^T$. The regression procedure is also applied to $T_z$ by training the distance MLP, but only utilises $\mu_{i}$, $w_i$, $h_i$, and $y_i$. Once $T_z$ and $(c_x,c_y)^T$ are determined, $T_x$ and $T_y$ are calculated using the pinhole camera model (Eq.~\ref{eq:translation}).
\begin{equation} \label{eq:translation}
  \begin{bmatrix}
    T_x\\
    T_y
  \end{bmatrix}
  = \begin{bmatrix}
    (c_x-p_x)\frac{T_z}{f_x}\\
    (c_y-p_y)\frac{T_z}{f_y}
  \end{bmatrix}
\end{equation}
where $f_x$ and $f_y$ denote the focal lengths, $(p_x,p_y)^T$ is the principal point, and all these parameters can be obtained from camera calibration.

\section{Experiments}
\label{sec:experiments}

The CVAM-Pose method is benchmarked in two aspects. The first involves conducting a series of ablation tests to determine favourable configurations of the method. The second way is to follow the evaluation methodologies proposed in the BOP challenges~\cite{hodavn2020bop,sundermeyer2023bop}.

\subsection{Experimental Setup}
\label{subsec:implementation}

\noindent\textbf{Data}\hspace{10mm}All the experiments are conducted using the Linemod-Occluded dataset~\cite{brachmann2014learning,lmodata}, as it presents a wide range of challenging scenarios, such as texture-less objects with significant occlusion and background clutter. To facilitate a fair comparison with methods participating in the BOP challenges, the same training and test data are employed. The physically based rendering (PBR) images~\cite{hinterstoisser2013model,hodavn2019photorealistic,hodavn2020bop,pbrdata} are used for training, and the BOP version test set is chosen for evaluation.

\vspace{12pt}

\noindent\textbf{Evaluation pipeline}\hspace{10mm}All the results, including those from the ablation tests and the main evaluation, are reported using the metrics specified in the BOP challenges: $\mathrm{VSD}$, $\mathrm{MSSD}$, and $\mathrm{MSPD}$~\cite{hodavn2020bop}. The overall performance score, $\mathrm{AR}$, is calculated based on the average recall of these three metrics, defined as $\mathrm{AR=(AR_{VSD}+AR_{MSSD}+AR_{MSPD})/3}$. 

\subsection{Ablation Study}
\label{subsec:ablation}

To obtain effective configurations of the CVAM-Pose method, extensive ablation tests are conducted using the BOP version of the Linemod-Occluded dataset~\cite{brachmann2014learning,lmodata,hodavn2020bop,pbrdata}. These tests include evaluations of the adapted label embedding technique, the regularisation of the label-embedded CVAE network, and the dimensionality of the latent space.

\vspace{12pt}

\noindent\textbf{Label Embedding Technique}\hspace{10mm}The effectiveness of the adapted layer-wise one-hot encoding technique is assessed in both the CVAE network and the MLPs. The original CVAE network~\cite{sohn2015learning}, where the label conditions only exist in the initial layer in both the encoder and decoder, is trained under the same conditions as our proposed label-embedded network. The tests on MLPs involve determining whether the representations learnt from the label-embedded CVAE network can be effectively regressed without the labels.

\begin{table}[ht]
\centering
\small
\setlength\tabcolsep{4pt} 
\begin{tabular}{|c|c|c|c|}
\hline
Network & original CVAE & MLPs without labels & \textbf{Ours} \\
\hline
$\mathrm{AR_{VSD}}$ & 0.256 & 0.251 & 0.346 \\
\hline
$\mathrm{AR_{MSSD}}$ & 0.255 & 0.243 & 0.362 \\
\hline
$\mathrm{AR_{MSPD}}$ & 0.630 & 0.554 & 0.714 \\
\hline
$\mathrm{AR}$ & 0.380 & 0.349 & 0.475 \\
\hline
\end{tabular}
\caption{Ablation results on the adapted label embedding technique.}
\label{table:cvae6d ablation label results}
\end{table}

The results presented in Table~\ref{table:cvae6d ablation label results} demonstrate that the adapted label-embedding technique enhances the ability to learn and regress pose representations. The proposed label-embedded CVAE network yields the most promising results compared to the original CVAE and MLPs, showing improvements of 10\% and 13\% respectively in $\mathrm{AR}$. This improvement is attributed to its capability to abstract high-level features related to object pose within the latent space. In contrast, the original CVAE network primarily incorporates low-level features, which is less effective for learning distinct multi-object representations because, with the increasing depth of the network, the conditioned features introduced at the initial layers may not be evident in the latent space. Similarly, the MLPs benefit from the label-embedding technique that helps regress distinct poses for each object. However, it is observed that even in the absence of label conditions, the learnt representations still retain certain categorical information, leading to reasonable results.

\vspace{12pt}

\noindent\textbf{Latent Space Regularisation}\hspace{10mm}When training the CVAM-Pose, the weighting factor $\alpha$ of the KL regularisation term plays a crucial role in determining the smoothness of the latent space. To find a good balance that allows for both informative latent space and robust generalisation, the proposed CVAE network is trained with different values of $\alpha \in [0,0.1,0.5,1]$. The pose estimation results corresponding to these values are reported in Table~\ref{table:ablation test regularisation}.

\begin{table}[ht]
\centering
\small
\setlength\tabcolsep{4pt}
\begin{tabular}{|c|c|c|c|c|} 
\hline
Regularisation & $\alpha=0$ & $\bm{\alpha} = \mathbf{0.1}$ & $\alpha=0.5$ & $\alpha=1$  \\
\hline
$\mathrm{AR_{VSD}}$ & 0.316  & 0.346 & 0.319 & 0.336 \\
\hline
$\mathrm{AR_{MSSD}}$ & 0.340 & 0.362 & 0.352 & 0.353 \\
\hline
$\mathrm{AR_{MSPD}}$ & 0.712 & 0.714 & 0.686 & 0.685 \\
\hline
$\mathrm{AR}$ & 0.456 & 0.475 & 0.452 & 0.458 \\
\hline
\end{tabular}
\caption{Ablation results on the regularisation of the latent space.}
\label{table:ablation test regularisation}
\end{table}

Based on the reported results, an $\alpha$ value of $0.1$ is identified as most effective, providing sufficient regularisation of the latent space without overly constraining it. Specifically, when $\alpha=0$, the CVAE network lacks control over the assumed prior Gaussian distribution in the latent space, leading to unrestricted $\mu$, and $\sigma^2$ approaching $0$, which results in suboptimal performance in comparison to $\alpha=0.1$. Conversely, when $\alpha$ is set to $0$ or $1$, the latent space is greatly affected by the KL divergence, which seems to overly smooth the distribution. This excessive smoothing may cause a loss of critical pose-related information, as the network focuses on minimising KL divergence over retaining distinctive features of the input data.

\vspace{12pt}

\noindent\textbf{Dimensionality of the Latent Space}\hspace{10mm}The dimensionality of the latent space, denoted as $n$, determines the capacity of the proposed CVAE network to encapsulate information about objects. Previous research, such as that by~\citet{sundermeyer2020augmented}, has explored the effect of latent space dimensionality on pose estimation; however, their ablation tests were limited to $n\leq128$ and focused solely on single-object scenarios. This limitation prompts further investigation into the performance impact of $n>128$ across a broader range of objects, as a single object may not be sufficient to reflect the complexities of an entire dataset.

\begin{table}[ht]
\centering
\small
\setlength\tabcolsep{4pt}
\begin{tabular}{|c|c|c|c|c|c|c|} 
\hline
Dimensionality & $n=32$ & $n=64$ & $n=128$ & $\bm{n} = \mathbf{256}$ & $n=512$ & $n=1024$ \\
\hline
$\mathrm{AR_{VSD}}$ & 0.226 & 0.301 & 0.283 & 0.346 & 0.306 & 0.263 \\
\hline
$\mathrm{AR_{MSSD}}$ & 0.224 & 0.318 & 0.303 & 0.362 & 0.317 & 0.284 \\
\hline
$\mathrm{AR_{MSPD}}$ & 0.528 & 0.654 & 0.681 & 0.714 & 0.706 & 0.695 \\
\hline
$\mathrm{AR}$ & 0.326 & 0.424 & 0.422 & 0.475 & 0.443 & 0.414 \\
\hline
\end{tabular}
\caption{Ablation results on the dimensionality of the latent space.}
\label{table:ablation test dimensionality}
\end{table}

To determine an effective size of the latent space for multi-object pose estimation, comprehensive experiments are conducted using all the Linemod-Occluded objects with dimensionalities set at $n\in[32,64,128,256,512,1024]$. Results, as detailed in Table~\ref{table:ablation test dimensionality}, show that the highest accuracy is observed at $n=256$. However, notably, the accuracy decreases at higher dimensionalities such as $512$ and $1024$, suggesting that an overly large latent space may not effectively contribute to pose encoding, and could potentially lead to diminished performance due to the CVAE network capturing too much variability in the latent space, overfitting to the specific training set.

\subsection{Main Results and Discussion}
\label{subsec:results}

\begin{figure}[ht]
\centering
\includegraphics[width=0.8\textwidth]{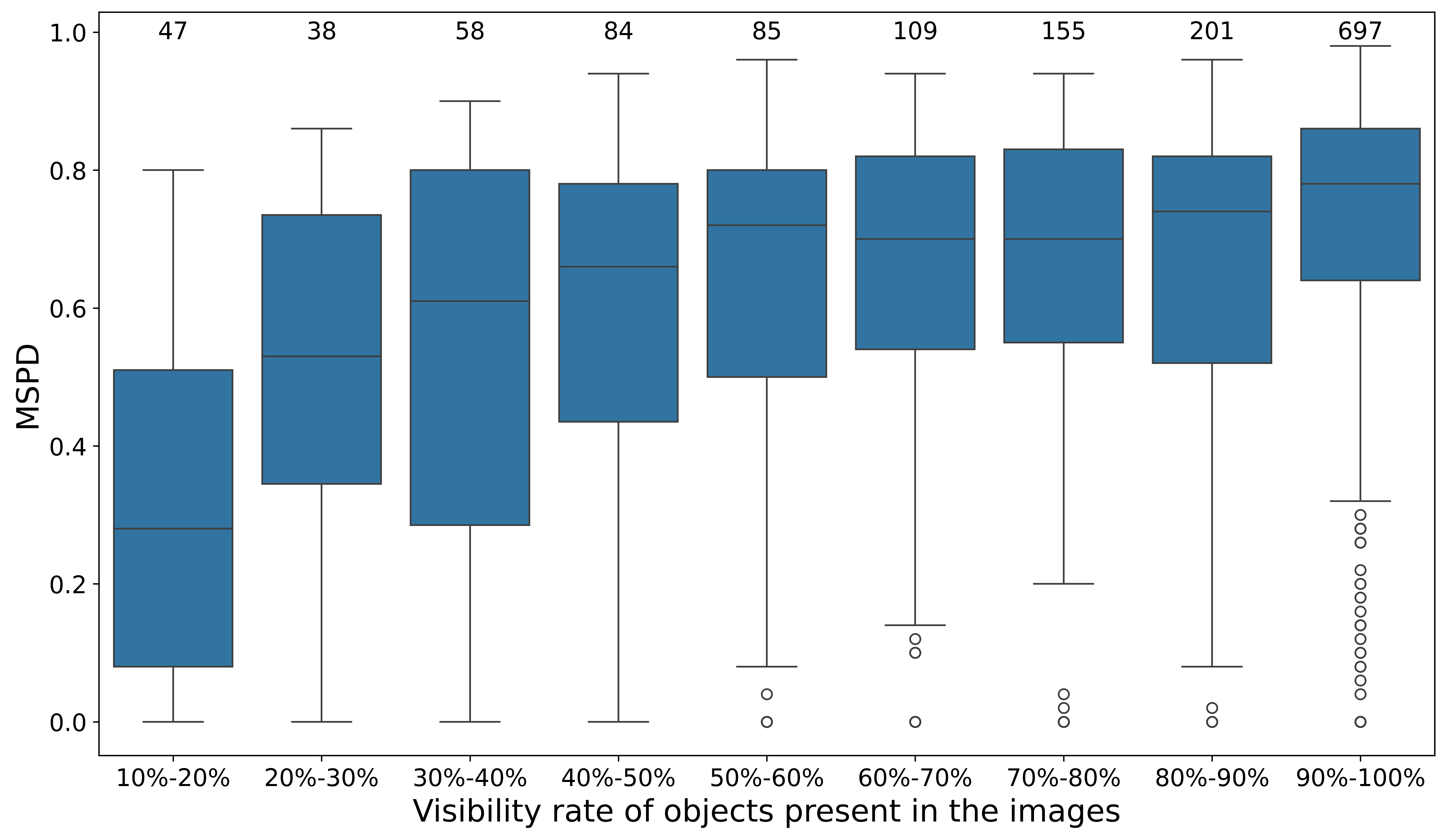}
\caption{Box plots of the MSPD metric as a function of the objects' visibility rates. The number of data instances for each rate is shown above each box. Please note that for better visualisation, the MSPD metric is calculated using thresholds ranging from $1$ to $50$ with a step of $1$, instead of using the thresholds (from $5$ to $50$ with a step of $5$) defined in the BOP challenges.}
\label{fig:occlusion boxplot}
\end{figure}

\noindent\textbf{Main Results}\hspace{10mm}The main results of the proposed CVAM-Pose method are reported in Table~\ref{table:lmo main results} and~\ref{table:lmo main results 2}, where it is compared against a variety of state-of-the-art methods on the BOP version of the Linemod-Occluded dataset. These methods are categorised based on the criteria outlined in Sec.~\ref{sec:related}, with the CVAM-Pose method classified within the latent representation category. Symbol "*" next to a method indicates that it employs one network per object class strategy. Additionally, we produce the box plots in Fig.~\ref{fig:occlusion boxplot}, which access how the visibility of objects (occlusion) in the scene images influences the pose estimation accuracy.

\begin{table}[ht]
\centering
\footnotesize
\setlength\tabcolsep{4pt} 
\begin{tabular}{|c|c|c|c|c|c|}
\hline
Method & \textbf{Ours} & CVML-Pose*~\cite{zhao2023cvml} & AAE*~\cite{sundermeyer2020augmented} & AAE-ICP*~\cite{sundermeyer2020augmented} & Multi-Path~\cite{sundermeyer2020multi} \\
\hline
$\mathrm{AR_{VSD}}$ & 0.346 & 0.312 & 0.090 & 0.208 & 0.150 \\
\hline
$\mathrm{AR_{MSSD}}$ & 0.362 & 0.338 & 0.095 & 0.218 & 0.153 \\
\hline
$\mathrm{AR_{MSPD}}$ & 0.714 & 0.706 & 0.254 & 0.285 & 0.346 \\
\hline
$\mathrm{AR}$ & 0.475 & 0.452 & 0.146 & 0.237 & 0.217 \\
\hline
\end{tabular}%
\caption{Comparison with latent representation methods.}
\label{table:lmo main results}
\end{table}

\begin{table}[ht]
\centering
\resizebox{\columnwidth}{!}{%
\begin{tabular}{|c|c|c|c|c|c|c|c|c|c|c|c|}
\hline
Method & \textbf{Ours} & DPOD~\cite{zakharov2019dpod} & Pix2Pose*~\cite{park2019pix2pose} & EPOS~\cite{hodan2020epos} & CDPN*~\cite{li2019cdpn} & PVNet*~\cite{peng2020pvnet} & CosyPose~\cite{labbe2020cosypose} & SurfEmb~\cite{haugaard2022surfemb} & ZebraPose*~\cite{su2022zebrapose} & GDR-Net*~\cite{wang2021gdr} \\
\hline
$\mathrm{AR_{VSD}}$ & 0.346 & 0.101 & 0.233 & 0.389 & 0.393 & 0.428 & 0.480 & 0.497 & 0.547 & 0.549 \\
\hline
$\mathrm{AR_{MSSD}}$ & 0.362 & 0.126 & 0.307 & 0.501 & 0.537 & 0.543 & 0.606 & 0.640 & 0.714 & 0.701 \\
\hline
$\mathrm{AR_{MSPD}}$ & 0.714 & 0.278 & 0.550 & 0.750 & 0.779 & 0.754 & 0.812 & 0.851 & 0.860 & 0.887 \\
\hline
$\mathrm{AR}$ & 0.475 & 0.169 & 0.363 & 0.547 & 0.569 & 0.575 & 0.633 & 0.663 & 0.707 & 0.713 \\
\hline
\end{tabular}%
}
\caption{Comparison with direct and indirect methods that are reliant on 3D models.}
\label{table:lmo main results 2}
\end{table}

Within the latent representation category, CVAM-Pose significantly outperforms methods such as AAE, AAE-ICP (AAE incorporates ICP refinement~\cite{besl1992method}), and Multi-Path (a multi-object AAE approach). For example, it is respectively 25\%, 14\%, and 20\% better when evaluated using the $\mathrm{AR_{VSD}}$ metric. The overall performance margins across the three metrics are substantial, with improvements of 33\%, 24\%, and 26\% in $\mathrm{AR}$ respectively. In comparison with methods from other categories, CVAM-Pose also surpasses some indirect methods like DPOD and Pix2Pose, by margins of 31\% and 11\%, respectively. Additionally, it achieves results comparable to EPOS, CDPN, and PVNet in specific metrics, e.g. $\mathrm{AR_{VSD}}$ and $\mathrm{AR_{MSPD}}$.

\vspace{12pt}

\noindent\textbf{Discussion}\hspace{10mm}The results on the challenging Linemod-Occluded dataset demonstrate that competitive performance for multi-object pose estimation can be achieved using a conditional generative model. The proposed label-embedded CVAE network with the continuous pose regression approach is more effective and accurate than other latent representation methods. Compared to the single-object methods, such as CVML-Pose, AAE, and AAE-ICP, the proposed label-embedded CVAE network captures regularised and robust representations for multiple objects. Unlike CVML-Pose, the adapted label-embedding technique avoids training multiple VAE networks, thereby enhancing training efficiency without diminishing the performance. Different from the Multi-Path method, which employs multiple decoder networks demanding significant GPU resources, our single-encoder-single-decoder architecture achieves higher pose accuracy. Furthermore, the continuous pose regression approach, as opposed to the LUT technique used in AAE, AAE-ICP, and Multi-Path, effectively avoids errors associated with pose discretisation during inference. Our continuous regression on the 2D projective centre and distance also mitigates the effects caused by incorrect detection of occluded objects.

Performance against occlusion is illustrated in Fig.~\ref{fig:occlusion boxplot}, which shows box plots quantifying the distribution of the MSPD metric ($\mathrm{AR_{MSPD}}$) across different visibility rates from 10\% to 100\%. It is evident that as visibility increases, the median of pose estimation accuracy improves and eventually achieves a value of $0.78$. Even under heavy (20\%-30\% visibility) or mild (50\%-60\% visibility) occlusions, our method still achieves reasonable results, indicating its robustness against challenging occlusion scenarios.

Compared to direct and indirect methods that rely on 3D models, the proposed CVAM-Pose method achieves higher results than some of them on the challenging occlusion data. This possibly suggests that approaches based on pixel-wise model correspondences, such as DPOD and Pix2Pose, suffer performance degradation due to an insufficient number of points available for assessing correspondences in heavily occluded scenes. In contrast, our proposed method benefits from reconstructing complete objects from partially obscured views, thereby robustly handling occlusions. 

For the results reported in the paper, the proposed CVAM-Pose method is trained with 8 different objects. Experiments with larger numbers of objects were also conducted but not reported. It was observed that increasing the number of objects in CVAM-Pose beyond 15 would lead to a decrease in pose estimation accuracy, without adjusting design parameters such as the size of the latent space.

Although the proposed CVAM-Pose method avoids training one network per object category, which enhances scalability and efficiency, it shows a certain gap in pose accuracy compared to leading methods like CosyPose, SurfEmb, and GDR-Net. These methods improve their accuracy through techniques such as iterative refinement using 3D model points (CosyPose), estimating continuous model correspondence distributions (SurfEmb), and combining pose regression with dense correspondences (GDR-Net). However, they all require precise 3D models for setting up 2D-3D correspondences or model point-based training. In contrast, the advantages of our method lie in addressing the 6-DoF pose estimation problem without relying on 3D models, depth measurements, and post-refinement processes, providing a novel solution in scenarios where such data are unavailable.

\section{Conclusion}
\label{sec:conclusion}

This paper addresses one of the key challenges in computer vision: finding multi-object 6-DoF poses from images captured by a perspective camera in real time (with fixed inference processing time of $0.02$s with CVAM-Pose run on RTX3090). The proposed method demonstrates that competitive performance can be achieved using only a single perspective image, without reliance on 3D models, depth measurements, or iterative post-refinement. In particular, the scalability of a single latent space can be expanded to multi-object representations without compromising pose accuracy. The main contributions of the reported research include the proposed use of a conditional generative model, the adapted label-embedding technique, the construction of a regularised and constrained latent space for multiple objects, and the continuous pose regression algorithms, which facilitate fast and accurate multi-object pose estimation.

\section{Acknowledgement}
\label{sec:acknowledgement}

Data access statement: The study reported in this paper has been supported by two existing openly available datasets, namely Linemod and Linemod-Occluded. Both datasets are available from \url{https://bop.felk.cvut.cz/datasets/}.

\newpage

\section*{Supplementary Materials}

We provide additional supplementary materials including:
\begin{enumerate}
  \item Further quantitative and qualitative analyses of our method on the BOP version of the Linemod-Occluded dataset~\cite{brachmann2014learning,lmodata,hodavn2020bop,pbrdata}.
  \item More information on network implementations.
\end{enumerate}

\section{Additional Results on Linemod-Occluded}
\label{sec:add results}

\subsection{Quantitative Results}
\label{subsec:quantitative}

\noindent\textbf{Pose Regression vs. LUT}\hspace{10mm}We conduct further ablation tests comparing the pose regression strategy used in our method to the lookup table (LUT) technique described in~\cite{sundermeyer2020augmented,sundermeyer2020multi}. The LUT technique assigns the rotation and projective distance from the most similar instance to the test instance, and utilises the centre of the bounding box as the 2D projective centre. This approach may lead to inaccuracies, particularly with heavily occluded objects or imprecise bounding boxes. In our analysis, the results for 3D rotation are reported using the $\mathrm{AR_{MSPD}}$ metric~\cite{hodavn2020bop}, while results for projective centre and distance are evaluated using the mean absolute error (MAE) metric. The choice of MAE over $\mathrm{AR_{MSPD}}$ is due to its parameter-free nature, which simplifies the interpretation of translational errors, as opposed to $\mathrm{AR_{MSPD}}$ that depends on predefined thresholds as outlined in~\cite{hodavn2020bop}.

\begin{table}[ht]
    \begin{minipage}{0.33\linewidth}
        \centering
        \begin{tabular}{|c|c|}
            \hline
            Rotation & $\mathrm{AR_{MSPD}} \uparrow $ \\
            \hline
            LUT & 0.666 \\
            \hline
            Ours & \textbf{0.714} \\
            \hline
        \end{tabular}
    \end{minipage}%
    \begin{minipage}{0.33\linewidth}
        \centering
        \begin{tabular}{|c|c|}
            \hline
            Centre & $\mathrm{MAE_{pixel}} \downarrow $ \\
            \hline
            LUT & 4.064 \\
            \hline
            Ours & \textbf{2.913} \\
            \hline
        \end{tabular}
    \end{minipage}%
    \begin{minipage}{0.33\linewidth}
        \centering
        \begin{tabular}{|c|c|}
            \hline
            Distance & $\mathrm{MAE_{mm}} \downarrow $ \\
            \hline
            LUT & 60.981 \\
            \hline
            Ours & \textbf{43.278} \\
            \hline
        \end{tabular}
    \end{minipage}
    \caption{Comparison between LUT and our regression method for the estimation of 3D rotation, 2D projective centre, and 2D projective distance.}
    \label{table:lmo compare our with lut}
\end{table}

\begin{figure}[ht]
\centering
\includegraphics[width=0.8\textwidth]{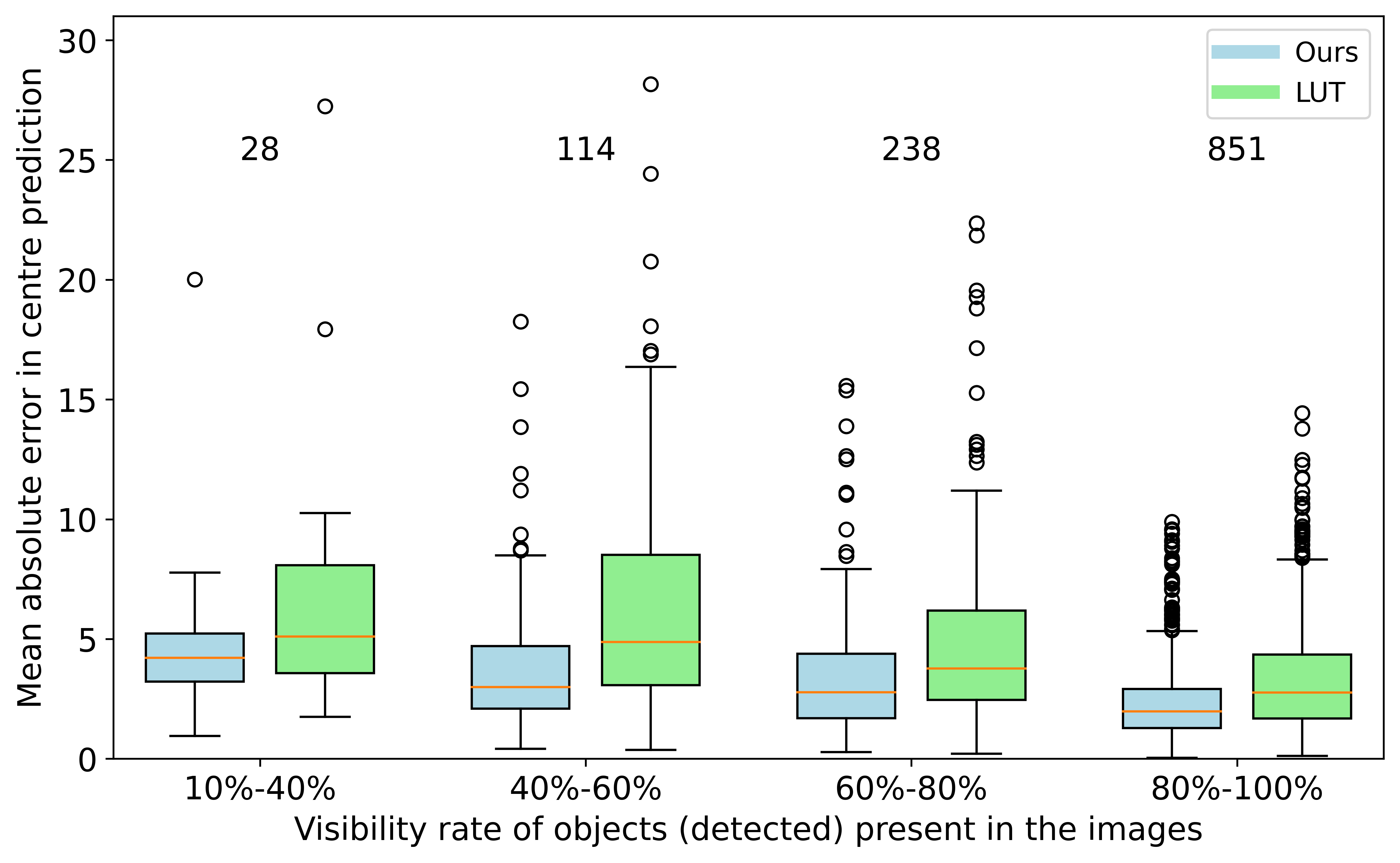}
\caption{Box plots of the $\mathrm{MAE_{pixel}}$ metric as a function of the objects' visibility rates. The number of data instances for each rate is shown above each pair of boxes.}
\label{fig:centre boxplot}
\end{figure}

As shown in Table~\ref{table:lmo compare our with lut}, our continuous pose regression strategy demonstrates better results than using the LUT technique in estimating 3D rotation, 2D projective centre, and 2D projective distance, e.g. our method achieves smaller errors in distance measurement (improved by approximately $2\%$ when computed in relation to the average object's distance in the test set). This can be attributed to the avoidance of the pose discretisation problem inherent in the LUT technique, particularly when the training data do not cover the entire $\mathrm{SO}(3)$. The performance of centre prediction is further illustrated in Fig.~\ref{fig:centre boxplot}, which presents box plots quantifying the distribution of errors ($\mathrm{MAE_{pixel}}$). It is evident that the median error in our method is consistently lower than that produced by the LUT technique across various visibility rates. The LUT method can also generate noticeable outlier errors in centre prediction, as high as $27$ pixels.

\vspace{12pt}

\noindent\textbf{Results on Individual Objects}\hspace{10mm}We also present additional results on individual objects from the Linemod-Occluded dataset~\cite{brachmann2014learning,lmodata} in Table~\ref{table:lmo individual results}. The average recall of a single object, $\mathrm{AR_{object}}$, is calculated from the average recall across the three metrics, $\mathrm{AR_{VSD}}$, $\mathrm{AR_{MSSD}}$, and $\mathrm{AR_{MSPD}}$~\cite{hodavn2020bop}. The average value, denoted as \textbf{Avg.}, shows the main results for the entire dataset as already reported in the paper.

\begin{table}[ht]
\centering
\setlength\tabcolsep{4pt} 
\begin{tabular}{|c|c|c|c|c|c|c|c|c|c|}
\hline
Object & ape & can & cat & driller & duck & eggbox & glue & holepuncher & \textbf{Avg.} \\
\hline
$\mathrm{AR_{VSD}}$ & 0.332 & 0.409 & 0.300 & 0.375 & 0.443 & 0.168 & 0.324 & 0.425 & 0.346 \\
\hline
$\mathrm{AR_{MSSD}}$ & 0.360 & 0.471 & 0.286 & 0.490 & 0.397 & 0.084 & 0.356 & 0.455 & 0.362 \\
\hline
$\mathrm{AR_{MSPD}}$ & 0.830 & 0.681 & 0.826 & 0.571 & 0.794 & 0.488 & 0.760 & 0.764 & 0.714 \\
\hline
$\mathrm{AR_{object}}$ & 0.507 & 0.520 & 0.471 & 0.479 & 0.545 & 0.247 & 0.480 & 0.548 & 0.475 \\
\hline
\end{tabular}
\caption{Results on the individual objects of the Linemod-Occluded dataset.}
\label{table:lmo individual results}
\end{table}

Among the three evaluation metrics, the $\mathrm{MSPD}$ metric demonstrates considerably higher accuracy than the others ($25\%$ higher on average). As explained in~\cite{hodavn2020bop}, this might be that the $\mathrm{MSPD}$ metric does not account for alignment along the optical axis, which is significant when evaluating on perspective images.

In terms of individual objects, the eggbox object exhibits lower accuracy than others (approximately $20\%$ in $\mathrm{AR_{object}}$), which might be associated with object symmetries, i.e. the pose ambiguity problem. To improve pose accuracy, especially for symmetrical objects, our method could be extended to estimate the distribution of potential poses through random sampling in the latent space, thereby better accommodating variances induced by object symmetries.

\subsection{Qualitative Results}
\label{subsec:qualitative}

\begin{figure}[p]
\centering
\includegraphics[width=0.8\textwidth]{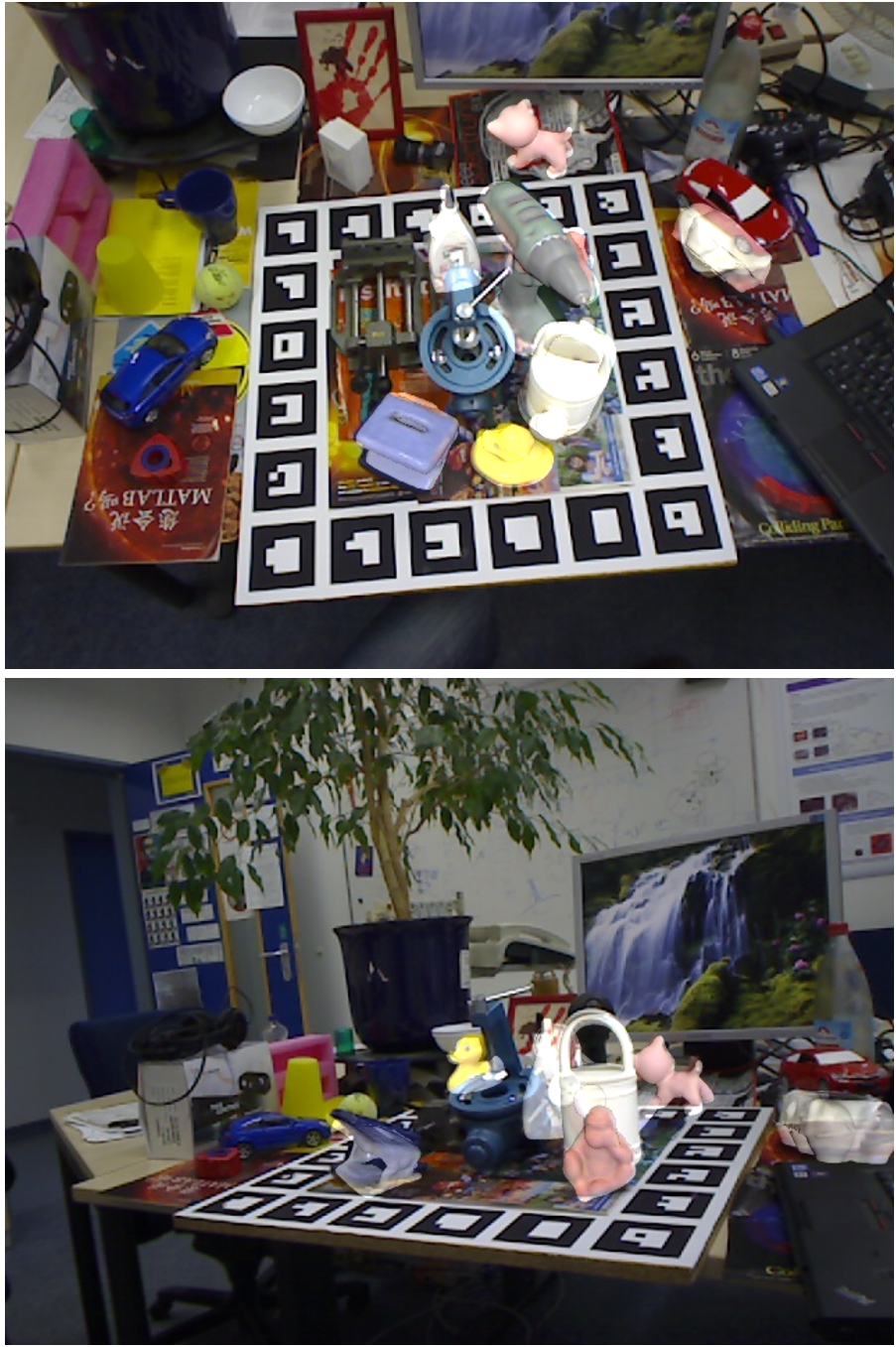}
\caption{Example visualisation of the estimated poses using CVAM-Pose. The rendering process uses the Pyrender software~\cite{pyrender}. The images of objects are taken from the BOP version of the Linemod-Occluded dataset~\cite{brachmann2014learning,lmodata,hodavn2020bop,pbrdata}.}
\label{fig:pose render}
\end{figure}

Fig.~\ref{fig:pose render} visualises pose estimation results on two randomly selected images from the Linemod-Occluded dataset, with poses estimated using CVAM-Pose. The target objects, including ape, cat, driller, duck, eggbox, glue, holepuncher, and iron, are rendered based on the estimated poses and reprojected onto the original test images. Correct estimations are represented by aligned reprojection masks, e.g. the cat object in the first image, while misaligned masks indicate incorrect estimations, e.g. the eggbox object in the first image.

\section{Implementation Details}
\label{sec:implementation details}

\noindent\textbf{Network Architecture}\hspace{10mm}The proposed label-embedded CVAE network employs an adapted ResNet-18~\cite{he2016deep} as the encoder, and a sequence of convolutional layers as the decoder. The ReLU activation function~\cite{nair2010rectified} is replaced with SiLU~\cite{elfwing2018sigmoid} to avoid the zero-gradient problem. The label-embedded MLP network consists of a series of fully connected layers with neurons $[256,128,64,32,16,out]$. Each hidden layer uses the SiLU activation and concatenates the one-hot encoded categorical labels with the output of the previous layer. The final output, $out$, varies depending on the regression task, such as $6$ neurons for regressing the continuous 6D rotation representation~\cite{zhou2019continuity}.

\vspace{12pt}

\noindent\textbf{Data Preprocessing}\hspace{10mm}The data preparation involves a crop-and-resize strategy proposed in~\cite{zhao2023cvml}. This strategy crops images of the target objects into a square shape from the scene image using the ground truth bounding box, with the square's size defined by the longer side of the box. The cropped images of objects are resized to $128\times128\times3$ using bicubic interpolation, which matches the input size of the proposed CVAE network. Images, where less than $10\%$ of the object’s area is visible, are excluded, based on the visibility criteria defined in~\cite{hodavn2020bop,sundermeyer2023bop}. Approximately $40k$ images per object are obtained, with $90\%$ designated for training and the remaining $10\%$ for validation. For test data preparation, the crop-and-resize strategy is also applied, using the detection bounding boxes provided by a pre-trained Mask-RCNN detector~\cite{he2017mask,labbe2020cosypose}.

\vspace{12pt}

\noindent\textbf{Training Parameters}\hspace{10mm}All experiments are implemented in PyTorch~\cite{paszke2019pytorch}. The label-embedded CVAE and MLP networks are trained using the AdamW optimiser~\cite{loshchilov2018decoupled} with parameters set as follows: $\beta_1=0.9$, $\beta_2=0.999$, $\epsilon=1e-8$, and $\lambda=0.01$. The initial learning rate is set to $1e-4$ for CVAE and $3e-3$ for MLPs, with scheduled reductions by a factor of $0.2$ when the validation loss does not improve over a “patience” period ($50$ epochs for CVAE, $500$ for MLPs). Training terminates when the lowest learning rate of $1e-6$ is reached, and no improvement in validation loss occurs for $N$ epochs ($N=50$ for CVAE and $N=1000$ for MLPs). The CVAE network is trained with a batch size of $128$, while MLPs process all inputs per batch. For reproducibility, all random seeds are fixed at $0$.

\newpage

\bibliography{egbib}

\end{document}